\documentclass{bmvc2k}
\usepackage{multirow}
\usepackage{booktabs}
\usepackage{amsmath,amssymb}
\usepackage{xcolor}
\usepackage{amsmath,amssymb,stmaryrd}   
\newcommand{\ie}{\emph{i.e}\bmvaOneDot}
\newcommand{\eg}{\emph{e.g}\bmvaOneDot}
\newcommand{\etal}{\emph{et al}\bmvaOneDot}

\title{FairFaceGAN: Fairness-aware Facial Image-to-Image Translation
\footnote{* Corresponding Author}
}

\addauthor{Sunhee Hwang}{sunny16@yonsei.ac.kr}{1}
\addauthor{Sungho Park}{qkrtjdgh18@yonsei.ac.kr}{1}
\addauthor{Dohyung Kim}{dohkim02@yonsei.ac.kr}{1}
\addauthor{Mirae Do}{wwwdo109@yonsei.ac.kr}{1}
\addauthor{Hyeran Byun*}{hrbyun@yonsei.ac.kr}{1}

\addinstitution{
 Department of Computer Science\\
 Yonsei University\\
 Seoul, Republic of Korea
}

\runninghead{Hwang et al.}{Fairness-aware Facial Image-to-Image Translation}

\def\eg{\emph{e.g}\bmvaOneDot}

\def\etal{\emph{et al}\bmvaOneDot}

\begin{document}

\maketitle
\begin{abstract}
In this paper, we introduce FairFaceGAN, a fairness-aware facial Image-to-Image translation model, mitigating the problem of unwanted translation in protected attributes (\eg, gender, age, race) during facial attributes editing.
Unlike existing models, FairFaceGAN learns fair representations with two separate latents - one related to the target attributes to translate, and the other unrelated to them.
This strategy enables FairFaceGAN to separate the information about protected attributes and that of target attributes. It also prevents unwanted translation in protected attributes while target attributes editing.
To evaluate the degree of fairness, we perform two types of experiments on CelebA dataset.
First, we compare the fairness-aware classification performances when augmenting data by existing image translation methods and FairFaceGAN respectively.
Moreover, we propose a new fairness metric, namely \textit{Fréchet Protected Attribute Distance} (FPAD), which measures how well protected attributes are preserved.
Experimental results demonstrate that FairFaceGAN shows consistent improvements in terms of fairness over the existing image translation models.
Further, we also evaluate image translation performances, where FairFaceGAN shows competitive results, compared to those of existing methods.

% Besides, our translation model can be trained without additional annotations for protected attributes, by employing a pre-trained classification network.

\end{abstract}

%-------------------------------------------------------------------------
\section{Introduction}
\label{sec:intro}
% AI intro & bias problem
Artificial Intelligence (AI) systems have achieved remarkable success in a broad range of research fields such as computer vision, natural language processing, and audio analysis. However, outputs of the AI systems could be biased since they heavily rely on human-collected datasets which may contain ethically sensitive stereotypes~\cite{geiger2020garbage}. Research and articles indicated that several AI systems yielded unfair results with respect to protected attributes such as gender, age, or race~\cite{propublica,sr_bias,unfair_od1,unfair_od2,unfair_od3,womanalso,balanced}. 
This is a critical problem to computer vision systems, which have already been deployed in diverse real world applications without adjusting demographic disparities.
For example, PULSE algorithm~\cite{pulse}, taking low-resolution faces into high-resolution images, tends to produce racially biased results, \ie, white skin, blue eyes, and brown hair, regardless of input images~\cite{sr_bias}.
%In the original image Obama has dark skin, black hair, and brown eyes, but the result is, instead, someone that has white skin, blue eyes, and brown hair.
%super resolution~\cite{sr_bias}
%For instance, ProPublica~\cite{propublica} pointed out that COMPAS, an AI based court system, was biased with respect to ethnicity. Although fairness and transparency are essential in the court system, it is more likely to predict African as a future criminal than other races. 
Accordingly, in order to resolve the societal bias problem, researchers have directed their attention on developing fair computer vision models~\cite{balanced, discoverfair, lnl, womanalso, manalso,fair1,fair2}. 
%Hence, the studies on fairness in computer vision are being carried out in a wide range of AI fields.

% fairness intro 

\begin{figure}[t]
\begin{tabular}{cc}
\includegraphics[width=6cm]{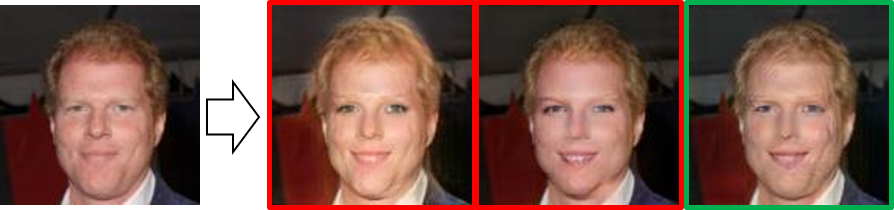}&
\includegraphics[width=6cm]{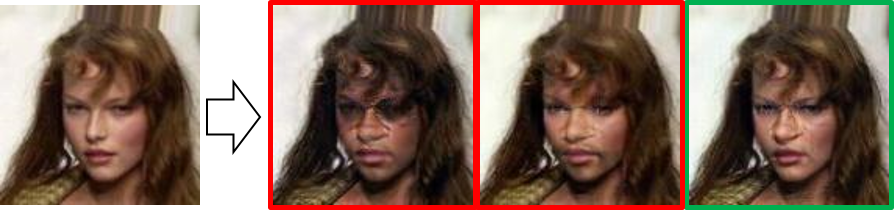}\\
 $+$ Attractive (\textcolor{red}{Gender changed})& $+$ Big Nose (\textcolor{red}{Race changed})\\
\includegraphics[width=6cm]{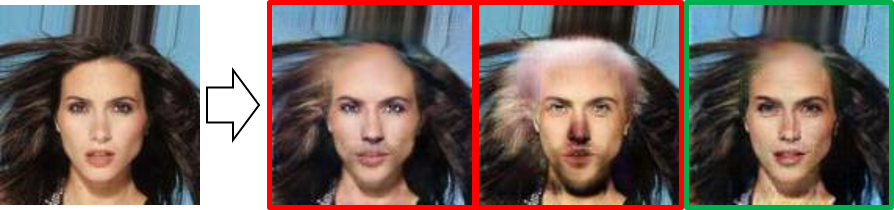}&
\includegraphics[width=6cm]{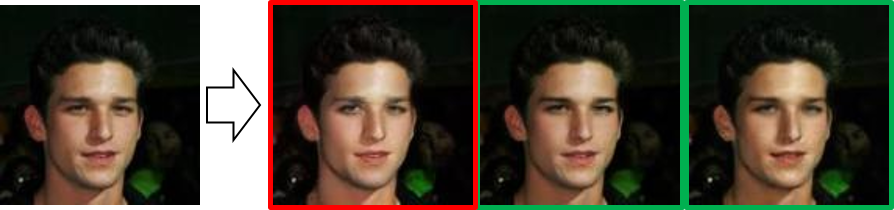}\\
 $+$ Bald (\textcolor{red}{Gender changed})& $-$ Bags Under Eyes (\textcolor{red}{Age changed})
\end{tabular}
\caption{Image translation results on CelebA dataset~\cite{celeba}. For each example, we present four facial images, which are an input image and the results of StarGAN, FixedPointGAN, and FairFaceGAN (ours), respectively (from left to right). $+$ and $-$ denote adding and removing the attribute of the input image, respectively. While \textcolor{red}{\textit{Red}} boxes indicate the occurrence of unwanted translation of protected attributes, \textcolor{green}{\textit{Green}} boxes denote the preservation of protected attributes. Best viewed in color.}
\label{fig:intro}
\end{figure}
%%%%%%%%%%%%%%%%%%%%%%%%%%%%%%%%%%%%%%%%%%%%%%%%%%%%

In this paper, we aim to improve fairness in Image-to-Image translation of facial attributes, whose goal is to edit attributes of input images. 
Even though recent methods based on Generative Adversarial Networks (GANs)~\cite{gan} succeeded in synthesizing realistic facial images while translating attributes fairly,
%Even though recent methods succeeded in synthesizing realistic facial images with well translated attributes based on Generative Adversarial Networks (GANs)~\cite{gan}, 
they might contain unintended discriminative factors.
% correlated to the target domain (\eg, protected attributes). 
In Figure~\ref{fig:intro}, we present several examples of discriminatory translation results. While translating of target attributes, existing facial attribute editing models~\cite{star,fp} unintendedly modify protected attributes (\ie, gender, age, race) as well.
% since the training datasets are biased to certain demographic groups.
%it produces images of the majority group in the training dataset.
%Inspired by a few attempts addressing the problem of unfair translation~\cite{fair1,fair2}, 

To address this problem, we propose a fairness-aware Image-to-Image translation model, namely FairFaceGAN, which maps input images into target domains while preserving protected attributes.
In specific, we introduce a new fair representation learning method that learns two separate latent spaces with different objectives: (i) one is for mapping target attributes adequately; (ii) the other is for preserving information of protected attributes.
By employing two decoupled latent spaces, FairFaceGAN successfully prevents unwanted translation during editing target attributes, as shown in the last column of each example of Figure~\ref{fig:intro}.
We note that our method can be easily extended to the case of multiple protected attributes as it separates target attributed-related information from the rest.
Moreover, another merit of FairFaceGAN is that it does not require protected attribute annotations.
Instead, we exploit knowledge related to protected attributes from a pre-trained classification model. 
We believe that this will largely benefit the application of our method especially in the circumstance where protected attribute labels are not acquirable. 

% Through extensive evaluation on CelebA dataset~\cite{celeba},
To compare FairFaceGAN with existing image translation models in terms of fairness, we design and perform two kinds of experiments.
Specifically, for the first one, we measure how the fairness-aware classification performances are improved when the biased training dataset is augmented by previous translation models and ours respectively.
For this, we use standard fairness metrics, \ie, Equality of Opportunity~\cite{eqopp} and Equalized Odds~\cite{odds}.
For the second one, we propose a new fairness metric, \textit{Fréchet Protected Attribute Distance} (FPAD), inspired by Fréchet Inception Distance (FID)~\cite{fid}, to evaluate the protected attribute preservation ability of image translation models.
On the both types of experiments, FairFaceGAN shows consistently fairer results over the existing image translation methods.
Also, we provide comparisons on the standard image translation metrics, \ie, FID and Kernel Inception Distance (KID), where FairFaceGAN achieves comparable results to the other models.

Our main contributions can be summarized as follows:
\begin{itemize}
    \item We introduce FairFaceGAN that maps input images into target domain in a fair way with respect to multiple protected attributes. 
    \item To reduce the correlation between protected and target attributes in the mapping, we propose to learn two separate representations with different objectives: target attributes mapping and protected attribute preservation.
    \item To achieve fairness, we present a knowledge transfer technique for fair translation on the target dataset. It enables our model to mitigate bias related to multiple protected attributes even for the case where annotations for protected attributes are unavailable.
    \item Through the extensive experiments on CelebA, we demonstrate that FairFaceGAN produces the fairest results in terms of Equality of Opportunity, Equalized Odds, and the proposed FPAD over existing Image-to-Image translation models.

\end{itemize}

\section{Related Work}

\subsection{Fairness in Computer Vision}
In recent years, fairness in computer vision has become a popular research topic. Among various types of fair methods, we briefly introduce two approaches to mitigate bias problems in visual recognition tasks: (1) Reorganizing a biased dataset to the fair dataset (Pre-processing), and (2) Reducing bias through model architecture or algorithm (In-processing). 

\paragraph{Pre-processing.} 
Sattigeri \etal~\cite{ibmfairnessgan} proposed a fair data generating method based on GANs. They are trained on a biased dataset and generate new data which are fair in terms of the protected attributes. The generated data is utilized to train a fairness-aware face attribute classification model. Quadrianto \etal~\cite{discoverfair} introduced a data-to-data translation method that transforms an original biased dataset into a new fair dataset. 
In this paper, we also address fairness in the image classification task by generating fair dataset using our FairFaceGAN.

\paragraph{In-processing.} 
Zheng \etal~\cite{cvpr_vae} proposed a disentangling method that splits feature representation into the two subspaces, one relevant to target labels and the irrelevant one. 
Similarly, FFVAE~\cite{ffr} aim to represent protected attribute related information and the rest. 
Park \etal~\cite{readme} proposed a fair disentangling method for representing target, protected attribute, and mutual information of both.
%split latent variables into three independent subspaces: target attribute relevant, protected attribute relevant, and the mutual information relevant spaces.
Unlike above, Wang \etal~\cite{balanced} proposed an adversarial approach to reduce gender bias in a visual recognition model. 
While, most existing methods take into account a binary protected attribute despite the diversity of demographic groups. 
In contrast, we introduce a fair method that eliminates multiple protected attributes related biases in computer vision models.
%Burns \etal~\cite{womanalso} propose a fairness-aware image captioning method that obstructs to generate gender-biased words. However, fairness in deep clustering
%has not been well addressed so far.

\subsection{Image-to-Image Translation} 
The main goal of Image-to-Image translation task is to learn how to map images from a source domain into images of a target domain. The methods based on Conditional Generative Adversarial Networks (CGANs)~\cite{cgan, pix2pix} have shown a great success with pixel-wise paired datasets in super-resolution~\cite{sr1}, image in-painting~\cite{inpainting1}, image restoration~\cite{derain}, and image segmentation~\cite{segment1}. In addition, Cycle consistency adversarial networks (CycleGANs)~\cite{cycle} are introduced to learn a mapping between unpaired datasets. They train the Image-to-Image translation models in an unsupervised manner. Moreover, Choi \etal~\cite{star} proposed StarGAN that reduces the computational cost of models based on CycleGAN. The unified and unsupervised Image-to-Image translation model learns a mapping between multiple domains effectively. However, we find out that the learned mapping is biased to protected attributes (See Figure~\ref{fig:intro}). There are some studies~\cite{fp,fair1,fair2} that prevent unwanted information translation during mapping. Although Siddiquee \etal~\cite{fp} proposed a FixedPointGAN that generates unchanged images in same-domain translation, it generates biased results in different-domain translation, a still remaining issue. In addition, fair representation methods by semantic constraints~\cite{fair1} and a disentangling method \cite{fair2} are developed. Inspired by~\cite{fair1,fair2}, we also aim to train a fairness-aware image translation model by proposing a fair representation learning method.

\begin{figure}[t]
\centering
\begin{tabular}{cc}
\includegraphics[width=10cm]{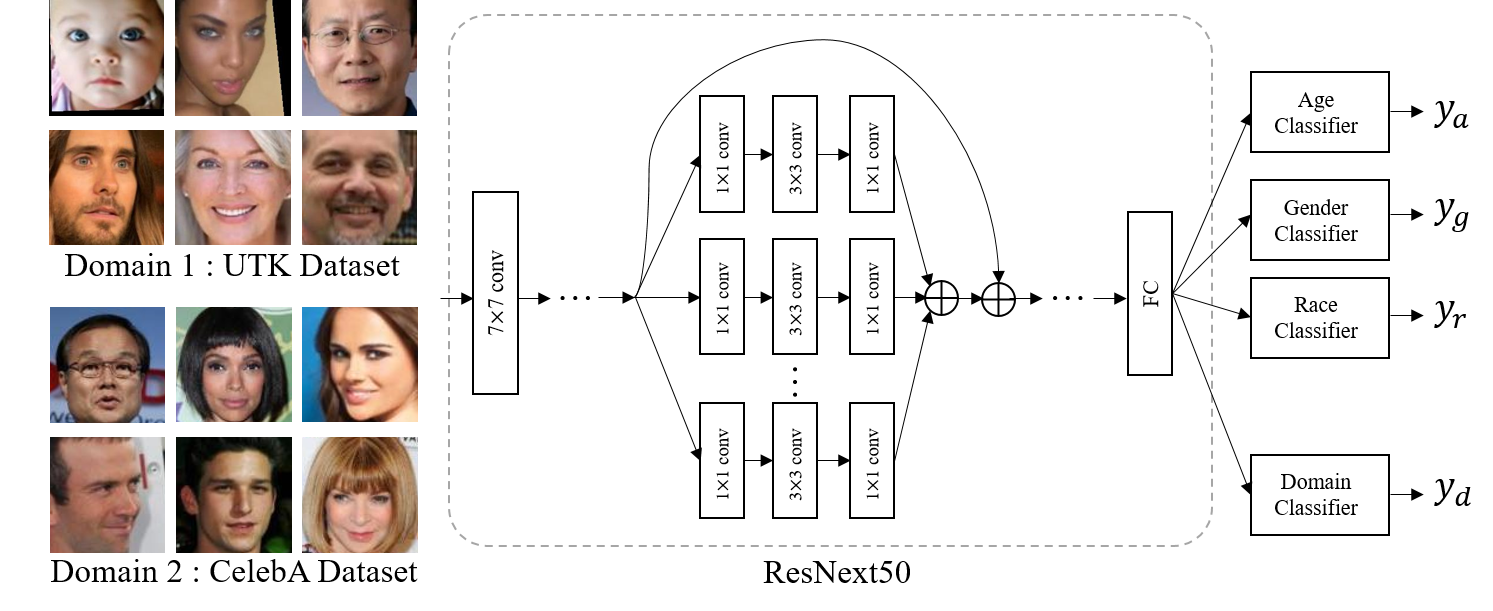}
\end{tabular}
\caption{The proposed Protected Attribute Classifier (PAC).}
\label{fig:pac}
\end{figure}

\section{Proposed Method}
In this work, we propose two modules: 1) Protected Attribute Classifier (PAC) module, which learns high-level features of multiple protected attributes. 2) FairFaceGAN, which is a fairness-aware facial Image-to-Image translation network to learn a fair mapping of the multiple facial attributes in the multi-domain. The main network for the fairness-aware Image-to-Image translation is FairFaceGAN and PAC module is introduced to train FairFaceGAN without protected attribute annotations. In this section, we explain the modules in sequence.

\subsection{Protected Attribute Classifier (PAC)}
As illustrated in Figure~\ref{fig:pac}, PAC consists of two branches: one is for predicting protected attributes (gender $y_g$, age $y_a$, race $y_r$) and the other is for predicting the domain labels $y_d$. The encoder of PAC with a number of convolutional layers is shared by the two branches and followed by task-specific fully connected layers: $f_g$ (gender classifier), $f_a$ (age classifier), $f_r$ (race classifier), and $f_d$ (domain classifier). We define the objective function for PAC as follows:

\begin{equation}
\mathcal{L}_{PA}= \mathcal{L}_{ce}( y_g | f_g( h ) ) + \mathcal{L}_{ce}( y_a | f_a( h ) )  + \mathcal{L}_{ce}( y_r | f_r( h ) ),	
\label{l:pa}
\end{equation}

\noindent where $\mathcal{L}_{ce}$ and $h$ respectively denote a cross-entropy loss and a flattened feature of the last layer from the shared encoder.

In addition, to transfer knowledge related to protected attributes from the learned PAC into the FairFaceGAN, we train a discriminator to fail classification on source domain (UTK dataset~\cite{utk}) and target domain(CelebA dataset~\cite{celeba}) using a gradient reversal layer like DANN~\cite{dann} since the representation of PAC and FairFaceGAN are trained on different domains. To do so, we optimize the domain adversarial loss as follows:

\begin{equation}
\mathcal{L}_{PAC}= \mathcal{L}_{PA} - \lambda \mathcal{L}_{ce}( y_d | fc_d( f(x) ) ).
\label{l:adv}
\end{equation}

\paragraph{Optimization} We use Adam optimizer with a learning rate of 0.001 and a batch size of 128. The PAC was optimized before ten epochs on a single 1080Ti GPU.

\begin{figure}[t]
\centering
\begin{tabular}{cc}
\includegraphics[width=12cm]{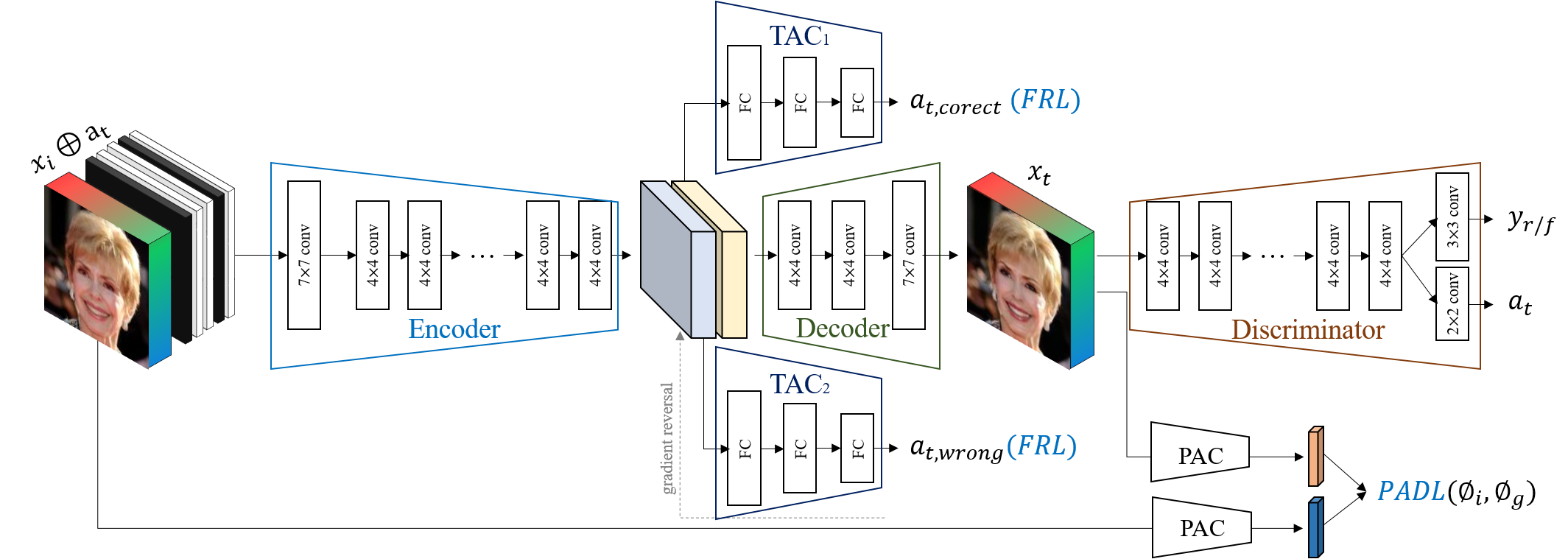}
\end{tabular}
\caption{An overview of the proposed FairFaceGAN framework, which consists of Encoder-Decoder Generator, Discriminator, and Target Attribute Classifiers (TACs). Given an image $x_i$ and target attribute $a_t$, we learn the model fairly work on protected attributes with Fair Representation Loss (FRL) and Protected Attribute Distance Loss (PADL) to generate $x_t$.}
\label{fig:ffg}
\end{figure}

\subsection{FairFaceGAN}
FairFaceGAN aims to map input images into target facial attributes using a unified generator. As shown in Figure~\ref{fig:ffg}, it contains four components: one encoder-decoder generator, two target attribute classifiers (TACs), and one discriminator.

Given an input image $x_i$ and a target attribute vector $a_t$, we first depth-wisely concatenate both of them. Then the combined data is fed into the encoder to represent two latent spaces. One is for target attributes and the other is for the rest information. The two features are then concatenated and used as an input of our decoder for generating a fair image $x_t$ with target attributes. % Lastly, the decoder generates an image $x_t$ with target attributes from the encoded features. 

\paragraph{Auxiliary Classifier Generative Adversarial Network Loss.}
We train FairFaceGAN with an adversarial loss to generate images to be realistic. In addition, we add an auxiliary classification layer on the top of the discriminator to distinguish the target attributes of the input image ($a_i$) and the generated image ($a_t$). The adversarial loss with the auxiliary classifier is defined as follows:

\begin{equation}
\label{l:ac}
\begin{split}
\underset{\theta_{G}}{min}~\underset{\theta_{D}}{max}\mathcal{L}_{acgan} =& \mathbb{E}_{x_i}[\log D(x_i)]+\mathbb{E}_{x_t}[\log (1-D(x_t))] \\ 
&- \mathbb{E}_{x_i,a_i}[\log p_{\theta_{D}}(a_i|x_i)] - \mathbb{E}_{x_t,a_t}[\log p_{\theta_{D}}(a_t|x_t)].
\end{split}
\end{equation}

\paragraph{Reconstruction Loss.}
For the reconstruction, we use a cycle consistency loss~\cite{cycle} that guarantees the quality of generated images in the unsupervised manner. In addition, inspired by FixedPointGAN~\cite{fp}, we add an identity loss to make the generative model not transfer unnecessary regions in a same-domain translation. 

\begin{equation}
\mathcal{L}_{rec}= \mathbb{E}_{x,a}[\left \|G(\widetilde{x}_t, a_i) - x_i  \right \|_1] + \mathbb{E}_{x,a}[\left \|G(x_i, a_i) - x_i  \right \|_1].
\label{l:rec}
\end{equation}

\paragraph{Fair Representation Loss (FRL).}

During translating target attributes, the high correlation between target attributes and protected attributes causes unwanted protected attribute translation. To prevent it, we separate representation $h$ into target attribute translation ($h_{tr}$) and protected attribute preservation ($h_{tu}$) respectively.
To this end, we apply a fair representation loss defined as follows:
\begin{equation}
\underset{\theta_{TAC_1},\theta_{ENC}}{\min}~\underset{\theta_{TAC_2}}{\max}\mathcal{L}_{fp}= \mathbb{E}_{x_i}[-\log p_{\theta_{TAC_1}}(a_t|h_{tr}) + \log p_{\theta_{TAC_2}}(a_t|h_{tu})].
\label{l:frl}
\end{equation}

\paragraph{Protected Attribute Distance Loss (PADL).}
In addition, we propose protected attribute distance loss (PADL) minimizes the protected attribute feature distance between input images ($\phi_i$) and generated images ($\phi_g$). Since we do not have protected attribute labels in the target dataset, we instead utilize the semantic knowledge of protected attribute from the trained PAC to measure the distance. With Fair Representation Loss (FRL), it explicitly preserves protected attribute information in target attribute translation. The loss is defined as follows:

\begin{equation}
\mathcal{L}_{pad}= \mathbb{E}_{x}[\left\| \phi_i - \phi_g  \right \|_1].
\label{l:gan}
\end{equation}

\paragraph{Perceptual Loss.}
On top of that, the perceptual loss \cite{percept} is used to improve the quality of outputs. We select the same layers of \cite{percept} to measure not only the style loss between input images and reconstructed images but also the content loss between input images and generated images.

\paragraph{Optimization}
We use WGAN with gradient penalty \cite{wgp} and Adam for optimizing the parameters of our FairFaceGAN with $\beta_1=$0.5 and $\beta_2=$0.999. We note that the overall loss function is a weighted sum of all terms. The initial learning rate for both generator and discriminator is set to 0.0001, which is decayed every eight epochs. We obtained the best results before 20 epochs on two 1080-TI GPUs.

\section{Experiments} 

\subsection{Dataset}

\paragraph{PAC.} We train PAC on UTK Face~\cite{utk} and CelebA~\cite{celeba} datasets. CelebA dataset is utilized only for domain adversarial training and UTK Face dataset is leveraged for protected attribute (gender, race, and age) classification training as well as domain adversarial training. We randomly select 19,708, 2,000, and 2,000 images of UTK dataset for training, validation, and test, respectively, where 200,599 images of CelebA dataset are set to the domain adversarial training. All images are resized to 128 $\times$ 128. Results with ranges of age and race for the classification are shown in Table \ref{tab:pac}.

\paragraph{FairFaceGAN.} For training FairFaceGAN, we use only CelebA dataset without protected attribute annotation.
Instead, by transferring knowledge from pre-trained PAC on UTK dataset, we utilize the protected attribute related semantic information.
Training and test datasets are composed of 200,599 and 2,000 respectively. We pre-process all images by randomly cropping (178 $\times$ 178) and resizing into 128 $\times$ 128. The five target attributes (\textit{attractive}, \textit{blond hair}, \textit{bags under eyes}, \textit{bald}, \textit{big nose}) are selected manually. While we conduct both qualitative and quantitative evaluation for the \textit{gender} attribute, we only conduct qualitative evaluation for the age and race attributes since their labels are not included in CelebA dataset.

\begin{figure}[t]
\centering
\begin{tabular}{cc}
\includegraphics[width=12cm]{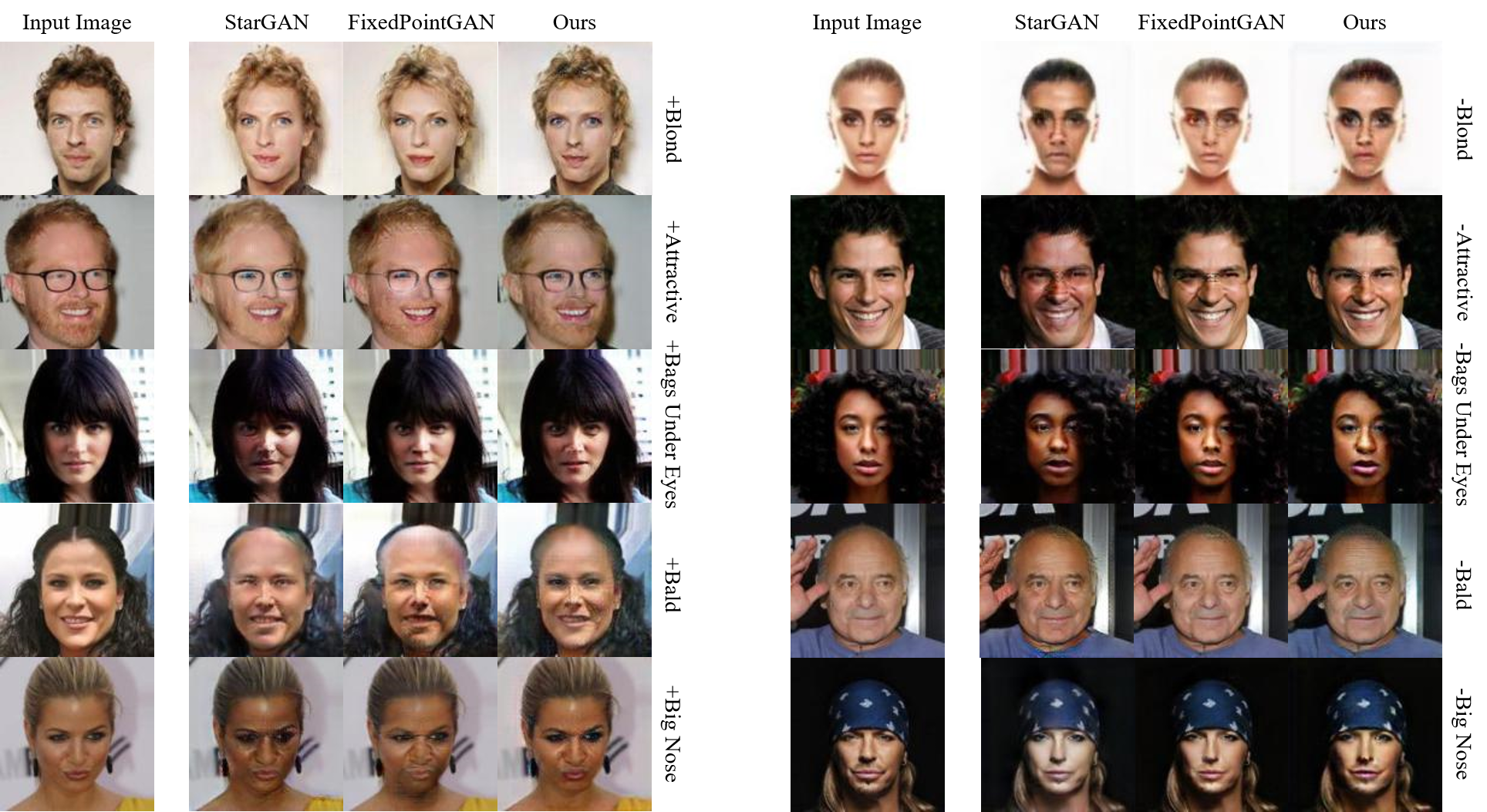}\\
\includegraphics[width=12cm]{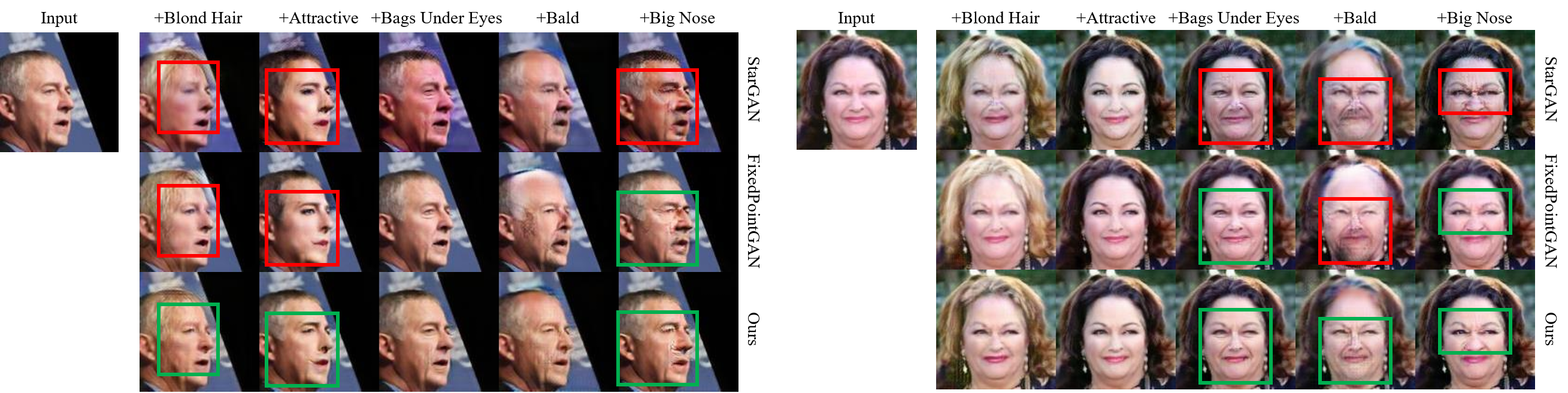}
\end{tabular}
\caption{Image-to-Image translation results compare to StarGAN~\cite{star} and FixedPointGAN~\cite{fp}. $+$ and $-$ denote the case of target attribute is added or removed.
\textcolor{red}{\textit{Red}} and \textcolor{green}{\textit{Green}} boxes indicate the discriminative outputs and fairly mapped results respectively.
}
%The set of top left and right denote examples of the target attribute is added and removed respectively. The set of Bottom left and right are examples of male and female images.}
\label{fig:comp}
\end{figure}

\begin{table}
\centering
\caption{Protected attribute classification accuracy on UTK dataset \cite{utk} (Source Only and DA). DA denotes results of the domain adversarial training.}
\label{tab:pac}
\begin{tabular}{lccc}
\toprule
Attribute {[}Label{]}                          & Source Only & DA & CelebA~\cite{celeba}\\
\midrule 
Gender {[}Male, Female{]}                      & 0.94        & 0.91 & 0.92             \\
Race {[}White, Black, Asian, Indian, Others{]} & 0.87        & 0.81  &  N/A           \\
Age {[}0$\sim$9, 10$\sim$19, … , 50+{]}        & 0.73        & 0.65   & N/A           \\
Domain Classification {[}UTK, CelebA{]}        & N/A         & 0.5     & N/A         \\\bottomrule 
\end{tabular}
\end{table}

\begin{table}
\centering
\caption{Quantitative comparison on CelebA dataset. f, p, and P indicate the usage of FRL, PADL, and Perceptual Loss. ACC, FID, and KID denote the average of target attribute classification accuracies, Fréshet Inception Distance~\cite{fid}, and Kernel Inception Distance ($\times$ 100)~\cite{kid}. }
\label{tab:attrcls}
\begin{tabular}{lcccccc}
\toprule
 \begin{tabular}[c]{@{}c@{}} \\ \end{tabular} & \begin{tabular}[c]{@{}c@{}}Star\\ GAN \cite{star}\end{tabular} & \begin{tabular}[c]{@{}c@{}}FixedPoint\\ GAN \cite{fp}\end{tabular} & \begin{tabular}[c]{@{}c@{}}Ours\\ (f)\end{tabular} & \begin{tabular}[c]{@{}c@{}}Ours \\ (p)\end{tabular} & \begin{tabular}[c]{@{}c@{}}Ours\\ (f+p)\end{tabular} & \begin{tabular}[c]{@{}c@{}}Ours\\ (f+p+P)\end{tabular} \\
\midrule
ACC $\uparrow$ & 92.07  & 91.01  & 90.55 & \textbf{92.11} & 89.71  & 90.66 
\\
FID $\downarrow$ & 10.23  & \textbf{6.91}  & 10.66 & 6.98 & 9.98  & 9.8 
\\
KID $\downarrow$ & 1.94$\pm$0.29 & 2.06$\pm$0.41 & 2.33$\pm$0.28  & \textbf{1.47}$\pm$0.35 & 2.13$\pm$0.3 & 1.89$\pm$\textbf{0.27}  
\\
\bottomrule
\end{tabular}
\end{table}

\begin{table}
\centering 
\caption{User study results.}
\label{tab:user}
\begin{tabular}{llll}
\toprule
         & StarGAN \cite{star}& FixedPointGAN \cite{fp}& Ours \\
         \midrule
Quality  & 30.78   & 20.97          & \textbf{48.25} \\
Fairness & 11.31    & 34.46          & \textbf{54.23}\\
\bottomrule
\end{tabular}
\end{table}

\begin{table}[!htb]
\caption{\textit{Fréshet Protected Attribute Distance} (FPAD) of generated images. BUE denotes \textit{Bags Under Eyes}. ($- \rightarrow +$) denotes without attribute into with attribute, and vice versa. }
\label{tab:fpad}
\begin{tabular}{llcccccc}
\toprule
\begin{tabular}[c]{@{}l@{}}Gender \\ (transform)\end{tabular}           & Attribute     & \begin{tabular}[c]{@{}c@{}}StarGAN\\ \cite{star}\end{tabular} & \begin{tabular}[c]{@{}c@{}}FixedPoint\\ GAN \cite{fp}\end{tabular} & \begin{tabular}[c]{@{}c@{}}Ours\\ (f)\end{tabular} & \begin{tabular}[c]{@{}c@{}}Ours\\ (p)\end{tabular} & \begin{tabular}[c]{@{}c@{}}Ours\\ (f+p)\end{tabular} & \begin{tabular}[c]{@{}c@{}}Ours\\ (f+p+P)\end{tabular} \\ \hline
\multirow{5}{*}{\begin{tabular}[c]{@{}l@{}}Male \\ ($- \rightarrow +$)\end{tabular}}   & BlondHair     & 56.32                                                & 24.55                                                    & 31.05                                              & 32.54                                              & \textbf{4.86}                                        & 5.63                                                   \\
                                                                       & Bald          & 11.68                                                & 11.90                                                    & 6.67                                               & 14.24                                              & \textbf{5.19}                                        & 8.30                                                   \\
                                                                       & BUE & 6.38                                                 & 2.60                                                     & 2.18                                               & 3.41                                               & \textbf{1.41}                                        & 3.44                                                   \\
                                                                       & BigNose       & 16.20                                                & 7.05                                                     & 4.62                                               & 9.99                                               & \textbf{1.51}                                        & 4.94                                                   \\
                                                                       & Attractive    & 11.32                                                & 3.49                                                     & 4.84                                               & 3.39                                               & \textbf{2.94}                                        & 3.79                                                   \\ \hline
\multirow{5}{*}{\begin{tabular}[c]{@{}l@{}}Male \\ ($+ \rightarrow -$)\end{tabular}}   & BlondHair     & 41.37                                                & 21.04                                                    & 20.11                                              & 32.01                                              & 9.96                                                 & \textbf{8.91}                                          \\
                                                                       & Bald          & 17.79                                                & 3.71                                                     & 3.97                                               & 8.51                                               & \textbf{2.19}                                        & 9.02                                                   \\
                                                                       & BUE & 21.29                                                & 6.87                                                     & 9.23                                               & 13.32                                              & \textbf{3.13}                                        & 5.23                                                   \\
                                                                       & BigNose       & 2.66                                                 & 2.02                                                     & 2.19                                               & 3.75                                               & \textbf{1.11}                                        & 1.63                                                   \\
                                                                       & Attractive    & 7.85                                                 & 4.43                                                     & 4.09                                               & 13.92                                              & \textbf{1.35}                                        & 6.7                                                    \\ \hline
\multirow{5}{*}{\begin{tabular}[c]{@{}l@{}}Female \\ ($- \rightarrow +$)\end{tabular}} & BlondHair     & 135.7                                                & 108.71                                                   & 72.98                                              & 104.13                                             & \textbf{4.75}                                        & 17.39                                                  \\
                                                                       & Bald          & 60.33                                                & 131.48                                                   & 22.18                                              & 57.83                                              & \textbf{21.00}                                       & 24.79                                                  \\
                                                                       & BUE & 3.25                                                 & 3.10                                                     & 1.71                                               & 3.08                                               & \textbf{1.55}                                        & 4.02                                                   \\
                                                                       & BigNose       & 22.42                                                & 12.18                                                    & 4.98                                               & 8.97                                               & \textbf{2.22}                                        & 3.47                                                   \\
                                                                       & Attractive    & 13.85                                                & 7.29                                                     & 6.17                                               & 3.05                                               & \textbf{2.78}                                        & 5.00                                                   \\ \hline
\multirow{4}{*}{\begin{tabular}[c]{@{}l@{}}Female \\ ($+ \rightarrow -$)\end{tabular}} & BlondHair     & 29.80                                                & 94.38                                                    & 35.49                                              & 55.39                                              & \textbf{5.17}                                        & 5.94                                                   \\
                                                                       & BUE & 6.06                                                 & 4.19                                                     & 9.57                                               & 4.42                                               & \textbf{2.29}                                        & 3.74                                                   \\
                                                                       & BigNose       & 5.77                                                 & 3.10                                                     & 4.95                                               & 4.5                                                & \textbf{2.18}                                        & 3.86                                                   \\
                                                                       & Attractive    & 22.79                                                & 13.36                                                    & 17.30                                              & 19.2                                               & \textbf{7.12}                                        & 11.70                                                  \\ \hline
\multicolumn{2}{l}{Average }                                                      & 25.94                                                & 24.50                                                    & 13.91                                              & 20.82                                              & \textbf{4.35}                                        & 7.24                                                   \\  \bottomrule
\end{tabular}
\end{table}

\begin{table}[!htb]
\centering 
\caption{Fair Classification Results. TPR, FPR, $Eq.Opp.$, and $Odds$ indicate Classification Accuracy, True Positive Rates, False Positive Rate, Equality of Opportunity~\cite{eqopp}, and Equalized Odds~\cite{odds}. $O$ and $G$ indicate the subset of original images in test dataset for the image translation model and the generator. Last three rows present results of data augmentation.}
\label{tab:acr}
\begin{tabular}{lcccccc}
\toprule
\multirow{2}{*}{Training Dataset} &  \multicolumn{2}{c}{Male} & \multicolumn{2}{c}{Female} & \multicolumn{2}{c}{Fairness Score} \\
     & TPR   & FPR   & TPR    & FPR  & $Eq.Opp.$ & $Odds$ \\
                         \midrule
$O$                        &64.10  & 18.40       & 86.36         & 49.00        & 22.26           & 26.43        \\
$G_{ours}(O)$                    & 79.49        & 29.45       & 90.40          & 53.00    & \textbf{10.92}           & 17.23   \\
\midrule
$O$+$G_{StarGAN}(O)$~\cite{star} &  64.10 & 15.34 & 91.41 & 43.00 & 27.31 & 27.49 \\
$O$+$G_{FixedPointGAN}(O)$~\cite{fp} & 56.41 & 19.63 & 87.88 & 42.00 & 31.47 & 26.92 \\

$O$+$G_{ours}(O)$         & 74.36        & 22.70   & 85.35         & 45.00        & \textbf{10.99}      & \textbf{16.65}    \\ \bottomrule   
\end{tabular}
\end{table}

\subsection{Evaluation}
\paragraph{Qualitative evaluation.}
As shown in Figure \ref{fig:comp}, FairFaceGAN generates better quality images compared to StarGAN \cite{star} and FixedPointGAN \cite{fp}. The models tend to change the skin color, add mustache on female images, apply makeup on male images, or make them aged, even though those are not the target attributes. Unlike their results, FairFaceGAN prevents the unwanted translation of protected attributes better.

\paragraph{Protected Attribute Classification.}
Table \ref{tab:pac} shows the protected attribute classification accuracy of PAC on UTK and CelebA datasets. We fine-tune the ImageNet \cite{imgnet} pre-trained ResNext50 \cite{resnext}, one of the state-of-the-art image classification networks. The result demonstrates that our PAC encodes representations informative to the protected attributes on both UTK and CelebA datasets.

\paragraph{Quantitative Comparisons.}
To compare quantitative results of generated images of ours and existing models, we measure the target attribute classification accuracy, Fréchet Inception Distance (FID)~\cite{fid}, and Kernel Inception Distance (KID)~\cite{kid}. In this experiment, we also conduct an ablation study of the proposed loss functions as follows: 1) Fair Representation Loss (FRL) only. 2) FRL and Protected Attribute Distance Loss (PADL). 3) FRL, PADL, and VGG Perceptual Loss. 
Firstly, to evaluate target attribute classification accuracies on the generated images, we re-train the ImageNet \cite{imgnet} pre-trained ResNext50 \cite{resnext} to classify the target attributes on CelebA dataset. As shown in Table \ref{tab:attrcls} (first row), the generated images from ours achieve the best result (92.11\%) over other models, where original testset achieves the accuracy of 88.88\%. We also measure FID and KID values to evaluate our model with standard metrics. As shown in Table \ref{tab:attrcls} (second and third rows), our model shows the best KID and competitive FID. Meanwhile, our final model shows slightly lower accuracy than others since there is a trade-off between fairness and the image generation ability~\cite{tradeoff1,tradeoff2}. Note that our goal focuses on improving fairness of the translation model.

\paragraph{User Study.}
We also present results of a user study to compare the fairness and visual quality of generated images of ours, StarGAN~\cite{star}, and FixedPointGAN~\cite{fp}. We randomly select 24 sets, four images per set (Input, Results of StarGAN, FixedPointGAN, and ours), and request 73 participants to choose the best produced (Quality) and the best protected attribute preserved (Fairness) images. As shown in Table \ref{tab:user}, our model achieves the best scores for both image quality and fairness.

\paragraph{Fréchet Protected Attribute Distance (FPAD).}
To evaluate the fairness of our proposed model, we propose a new metric FPAD inspired by FID \cite{fid}. We leverage our PAC model to extract a protected attribute feature and measure feature distance of input images $X_i$ and generated images $X_g$. We compute $||M_i - M_g||^2 + \text{Tr} (C_i + C_g - 2 (C_i C_g)^{1/2})$ in given ($M_i$, $C_i$) and ($M_g$, $C_g$) which are the mean and covariance of protected attribute features from $X_i$ and $X_g$. As shown in Table \ref{tab:fpad}, our model achieves the lowest FPAD compared to the prior models. In other words, our generative model best preserves the protected attributes during the mapping. Although there is a slight performance drop, we compensate it by applying the perceptual loss that improves visual quality of generated images.

\paragraph{Fair Classification.}
Furthermore, to evaluate our model using standard fairness metrics, we conduct an attractiveness classification task. We compare the performances when augmenting data by existing image translation models~\cite{fp,star} and FairFaceGAN respectively.
%by utilizing data augmentation, we evaluate how generated images improve fairness in facial attractiveness classification. 
For the evaluation, we leverage the two fairness metrics: Equality of Opportunity and Equalized Odds ($Eq.Opp.~=~|TPR_{male}-TPR_{female}|$, $Odds~=~\frac{1}{2}[|FPR_{male} - FPR_{female} | + | TPR_{male} - TPR_{female}|]$). Details are in our supplementary material. We fine-tune ImageNet pre-trained ResNext50 \cite{resnext} using the testset of FairFaceGAN, divided into 1,200 ($O$), 300, and 500 images for training, validation, and test, respectively. 
%For data augmentation, we modify the attractiveness of original images. 
As shown in Table~\ref{tab:acr}, we verify whether generated images of FairFaceGAN can be utilized for the classification model to be trained more fairly on gender compare to existing image translation models.
%Since the original dataset contains gender bias, existing image translation models tend to generate discriminative images.
% we compare face attractiveness classification results using original images ($O$) and performing data augmentation of generated images in terms of fairness scores:
% Our model achieves the fairest (lowest) results in terms of both Equality of Opportunity and Equalized Odds.
% \begin{equation}
% \begin{split}
% Eq.Opp. &= |TPR_{male} - TPR_{female}|, \\
% Odds &= \frac{1}{2}[|FPR_{male} - FPR_{female} | + | TPR_{male} - TPR_{female}|].
% \end{split}
% \end{equation}

%with respect to gender and increase $Eq.Opp$ and $Odds$. %Unlike the models, $Eq.Opp$ and $Odds$ are decreased when data augmentation is performed by our model with a graceful accuracy drop. 

\section{Conclusion}
In this paper, we introduced a novel fairness-aware facial Image-to-Image translation model to avoid the problem of translating unwanted attributes. Through Fair Representation Loss (FRL) and Protected Attribute Distance Loss (PADL), our model learns fair representations in terms of multiple protected attributes (age, gender, and race). To demonstrate the ability of FairFaceGAN, we conducted an extensive evaluation of image translation and fairness. Overall, our experimental results showed that FairFaceGAN is fairer in terms of Equality of Opportunity, Equalized Odds, and the proposed FPAD over the existing Image-to-Image translation models.

%By performing data augmentation by our model, our model achieved the fairest classification results in terms of Equality of Opportunity and Equalized Odds. In addition, we proposed a new measurement \textit{Fréchet Protected Attribute Distance (FPAD)} to measure how well our model preserves protected attributes. In overall, our experimental results showed that FairFaceGAN is fairer over the existing Image-to-Image translation models.

\paragraph{Acknowledgements.} 
%This work was supported by Institute for Information \& communications Technology Planning \& Evaluation (IITP) grant funded by the Korea government (MSIT) (No.2019-0-01396, Development of framework for analyzing, detecting, mitigating of bias in AI model and training data).
%This work was supported by Institute for Information \& communications Technology Planning \& Evaluation (IITP) grant funded by the Korea government (MSIT) (No.2020-0-01361, Artificial Intelligence Graduate School Program(YONSEI UNIVERSITY)).
This work was supported by Institute for Information \& communications
Technology Planning \& Evaluation (IITP) grant funded by the Korea government (MSIT) (Development of framework for analyzing, detecting, mitigating of bias in AI model and training data) under Grant 2019-0-01396 and (Artificial Intelligence Graduate School Program (YONSEI UNIVERSITY)) under Grant 2020-0-01361.

\noindent We thank \textbf{Pilhyeon Lee}, \textbf{Seogkyu Jeon}, and \textbf{Jijoong Kim} for the thorough reviews and the constructive feedback.

\bibliography{egbib}

\begin{thebibliography}{41}
\providecommand{\natexlab}[1]{#1}
\providecommand{\url}[1]{\texttt{#1}}
\expandafter\ifx\csname urlstyle\endcsname\relax
  \providecommand{\doi}[1]{doi: #1}\else
  \providecommand{\doi}{doi: \begingroup \urlstyle{rm}\Url}\fi

\bibitem[Anne~Hendricks et~al.(2018)Anne~Hendricks, Burns, Saenko, Darrell, and
  Rohrbach]{womanalso}
Lisa Anne~Hendricks, Kaylee Burns, Kate Saenko, Trevor Darrell, and Anna
  Rohrbach.
\newblock Women also snowboard: Overcoming bias in captioning models.
\newblock In \emph{Proceedings of the European Conference on Computer Vision
  (ECCV)}, pages 771--787, 2018.

\bibitem[Bi{\'n}kowski et~al.(2018)Bi{\'n}kowski, Sutherland, Arbel, and
  Gretton]{kid}
Miko{\l}aj Bi{\'n}kowski, Dougal~J Sutherland, Michael Arbel, and Arthur
  Gretton.
\newblock Demystifying mmd gans.
\newblock In \emph{International Conference on Learning Representations}, 2018.

\bibitem[Brandao(2019)]{unfair_od3}
Martim Brandao.
\newblock Age and gender bias in pedestrian detection algorithms.
\newblock In \emph{Proceedings of the IEEE Conference on Computer Vision and
  Pattern Recognition Workshops}, 2019.

\bibitem[Choi et~al.(2018)Choi, Choi, Kim, Ha, Kim, and Choo]{star}
Yunjey Choi, Minje Choi, Munyoung Kim, Jung-Woo Ha, Sunghun Kim, and Jaegul
  Choo.
\newblock Stargan: Unified generative adversarial networks for multi-domain
  image-to-image translation.
\newblock In \emph{Proceedings of the IEEE conference on computer vision and
  pattern recognition}, pages 8789--8797, 2018.

\bibitem[Creager et~al.(2019)Creager, Madras, Jacobsen, Weis, Swersky, Pitassi,
  and Zemel]{ffr}
Elliot Creager, David Madras, Jorn Jacobsen, Marissa Weis, Kevin~Jordan
  Swersky, Toniann Pitassi, and Richard Zemel.
\newblock Flexibly fair representation learning by disentanglement.
\newblock In \emph{Thirty-sixth International Conference on Machine Learn
  (ICML)}, 2019.

\bibitem[de~Vries et~al.(2019)de~Vries, Misra, Wang, and van~der
  Maaten]{unfair_od1}
Terrance de~Vries, Ishan Misra, Changhan Wang, and Laurens van~der Maaten.
\newblock Does object recognition work for everyone?
\newblock In \emph{Proceedings of the IEEE Conference on Computer Vision and
  Pattern Recognition Workshops}, 2019.

\bibitem[Dutta et~al.(2020)Dutta, Wei, Yueksel, Chen, Liu, and
  Varshney]{tradeoff2}
Sanghamitra Dutta, Dennis Wei, Hazar Yueksel, Pin-Yu Chen, Sijia Liu, and
  Kush~R Varshney.
\newblock Is there a trade-off between fairness and accuracy? a perspective
  using mismatched hypothesis testing.
\newblock In \emph{ICML 2020}, July 2020.

\bibitem[Ganin and Lempitsky(2015)]{dann}
Yaroslav Ganin and Victor Lempitsky.
\newblock Unsupervised domain adaptation by backpropagation.
\newblock In \emph{Proceedings of the 32nd International Conference on
  International Conference on Machine Learning - Volume 37}, ICML’15, page
  1180–1189. JMLR.org, 2015.

\bibitem[Geiger et~al.(2020)Geiger, Yu, Yang, Dai, Qiu, Tang, and
  Huang]{geiger2020garbage}
R~Stuart Geiger, Kevin Yu, Yanlai Yang, Mindy Dai, Jie Qiu, Rebekah Tang, and
  Jenny Huang.
\newblock Garbage in, garbage out? do machine learning application papers in
  social computing report where human-labeled training data comes from?
\newblock In \emph{Proceedings of the 2020 Conference on Fairness,
  Accountability, and Transparency}, pages 325--336, 2020.

\bibitem[Goodfellow et~al.(2014)Goodfellow, Pouget-Abadie, Mirza, Xu,
  Warde-Farley, Ozair, Courville, and Bengio]{gan}
Ian Goodfellow, Jean Pouget-Abadie, Mehdi Mirza, Bing Xu, David Warde-Farley,
  Sherjil Ozair, Aaron Courville, and Yoshua Bengio.
\newblock Generative adversarial nets.
\newblock In \emph{Advances in neural information processing systems}, pages
  2672--2680, 2014.

\bibitem[Gulrajani et~al.(2017)Gulrajani, Ahmed, Arjovsky, Dumoulin, and
  Courville]{wgp}
Ishaan Gulrajani, Faruk Ahmed, Martin Arjovsky, Vincent Dumoulin, and Aaron~C
  Courville.
\newblock Improved training of wasserstein gans.
\newblock In \emph{Advances in neural information processing systems}, pages
  5767--5777, 2017.

\bibitem[Hardt et~al.(2016)Hardt, Price, and Srebro]{eqopp}
Moritz Hardt, Eric Price, and Nati Srebro.
\newblock Equality of opportunity in supervised learning.
\newblock In \emph{Advances in neural information processing systems}, pages
  3315--3323, 2016.

\bibitem[Heusel et~al.(2017)Heusel, Ramsauer, Unterthiner, Nessler, and
  Hochreiter]{fid}
Martin Heusel, Hubert Ramsauer, Thomas Unterthiner, Bernhard Nessler, and Sepp
  Hochreiter.
\newblock Gans trained by a two time-scale update rule converge to a local nash
  equilibrium.
\newblock In \emph{Advances in neural information processing systems}, pages
  6626--6637, 2017.

\bibitem[Hwang and Byun(2020)]{fair2}
Sunhee Hwang and Hyeran Byun.
\newblock Unsupervised image-to-image translation via fair representation of
  gender bias.
\newblock In \emph{ICASSP 2020 - 2020 IEEE International Conference on
  Acoustics, Speech and Signal Processing (ICASSP)}, pages 1953--1957, 2020.

\bibitem[Isola et~al.(2017)Isola, Zhu, Zhou, and Efros]{pix2pix}
Phillip Isola, Jun-Yan Zhu, Tinghui Zhou, and Alexei~A Efros.
\newblock Image-to-image translation with conditional adversarial networks.
\newblock In \emph{Proceedings of the IEEE conference on computer vision and
  pattern recognition}, pages 1125--1134, 2017.

\bibitem[Johnson et~al.(2016)Johnson, Alahi, and Fei-Fei]{percept}
Justin Johnson, Alexandre Alahi, and Li~Fei-Fei.
\newblock Perceptual losses for real-time style transfer and super-resolution.
\newblock In \emph{European conference on computer vision}, pages 694--711.
  Springer, 2016.

\bibitem[Julia~Angwin and Kirchner()]{propublica}
Surya~Mattu Julia~Angwin, Jeff~Larson and Lauren Kirchner.
\newblock Machine bias.
\newblock URL
  \url{https://www.propublica.org/article/machine-bias-risk-assessments-in-criminal-sentencing}.

\bibitem[Kim et~al.(2019)Kim, Kim, Kim, Kim, and Kim]{lnl}
Byungju Kim, Hyunwoo Kim, Kyungsu Kim, Sungjin Kim, and Junmo Kim.
\newblock Learning not to learn: Training deep neural networks with biased
  data.
\newblock In \emph{Proceedings of the IEEE Conference on Computer Vision and
  Pattern Recognition}, pages 9012--9020, 2019.

\bibitem[Laradji et~al.(2019)Laradji, Vazquez, and Schmidt]{segment1}
Issam~H Laradji, David Vazquez, and Mark Schmidt.
\newblock Where are the masks: Instance segmentation with image-level
  supervision.
\newblock In \emph{British Machine Vision Conference (BMVC)}, Cardiff, UK,
  2019.

\bibitem[Liu et~al.(2015)Liu, Luo, Wang, and Tang]{celeba}
Ziwei Liu, Ping Luo, Xiaogang Wang, and Xiaoou Tang.
\newblock Deep learning face attributes in the wild.
\newblock In \emph{Proceedings of International Conference on Computer Vision
  (ICCV)}, December 2015.

\bibitem[Menon et~al.(2020)Menon, Damian, Hu, Ravi, and Rudin]{pulse}
Sachit Menon, Alexandru Damian, Shijia Hu, Nikhil Ravi, and Cynthia Rudin.
\newblock Pulse: Self-supervised photo upsampling via latent space exploration
  of generative models.
\newblock In \emph{Proceedings of the IEEE/CVF Conference on Computer Vision
  and Pattern Recognition}, pages 2437--2445, 2020.

\bibitem[Mirza and Osindero(2014)]{cgan}
Mehdi Mirza and Simon Osindero.
\newblock Conditional generative adversarial nets.
\newblock \emph{arXiv preprint arXiv:1411.1784}, 2014.

\bibitem[Park et~al.(2020)Park, Kim, Hwang, and Byun]{readme}
Sungho Park, Dohyung Kim, Sunhee Hwang, and Hyeran Byun.
\newblock Readme: Representation learning by fairness-aware disentangling
  method.
\newblock \emph{arXiv preprint arXiv:2007.03775}, 2020.

\bibitem[Quadrianto et~al.(2019)Quadrianto, Sharmanska, and
  Thomas]{discoverfair}
Novi Quadrianto, Viktoriia Sharmanska, and Oliver Thomas.
\newblock Discovering fair representations in the data domain.
\newblock In \emph{The IEEE Conference on Computer Vision and Pattern
  Recognition (CVPR)}, June 2019.

\bibitem[Russakovsky et~al.(2015)Russakovsky, Deng, Su, Krause, Satheesh, Ma,
  Huang, Karpathy, Khosla, Bernstein, et~al.]{imgnet}
Olga Russakovsky, Jia Deng, Hao Su, Jonathan Krause, Sanjeev Satheesh, Sean Ma,
  Zhiheng Huang, Andrej Karpathy, Aditya Khosla, Michael Bernstein, et~al.
\newblock Imagenet large scale visual recognition challenge.
\newblock \emph{International journal of computer vision}, 115\penalty0
  (3):\penalty0 211--252, 2015.

\bibitem[Sabato and Yom-Tov(2020)]{tradeoff1}
Sivan Sabato and Elad Yom-Tov.
\newblock Bounding the fairness and accuracy of classifiers from population
  statistics.
\newblock In \emph{ICML 2020}, July 2020.

\bibitem[Sattigeri et~al.(2019)Sattigeri, Hoffman, Chenthamarakshan, and
  Varshney]{ibmfairnessgan}
Prasanna Sattigeri, Samuel~C Hoffman, Vijil Chenthamarakshan, and Kush~R
  Varshney.
\newblock Fairness gan: Generating datasets with fairness properties using a
  generative adversarial network.
\newblock \emph{IBM Journal of Research and Development}, 63\penalty0
  (4/5):\penalty0 3--1, 2019.

\bibitem[Siddiquee et~al.(2019)Siddiquee, Zhou, Tajbakhsh, Feng, Gotway,
  Bengio, and Liang]{fp}
Md~Mahfuzur~Rahman Siddiquee, Zongwei Zhou, Nima Tajbakhsh, Ruibin Feng,
  Michael~B Gotway, Yoshua Bengio, and Jianming Liang.
\newblock Learning fixed points in generative adversarial networks: From
  image-to-image translation to disease detection and localization.
\newblock In \emph{Proceedings of the IEEE International Conference on Computer
  Vision}, pages 191--200, 2019.

\bibitem[Tsiminaki et~al.(2019)Tsiminaki, Dong, Oswald, and Pollefeys]{sr1}
Vagia Tsiminaki, Wei Dong, Martin~R. Oswald, and Marc Pollefeys.
\newblock Joint multi-view texture super-resolution and intrinsic
  decomposition.
\newblock In \emph{British Machine Vision Conference (BMVC)}, Cardiff, UK,
  2019.

\bibitem[Vincent(2020)]{sr_bias}
James Vincent.
\newblock What a machine learning tool that turns obama white can (and can’t)
  tell us about ai bias, 2020.
\newblock URL
  \url{https://www.theverge.com/21298762/face-depixelizer-ai-machine-learning-tool-pulse-stylegan-obama-bias}.

\bibitem[Wang et~al.(2019{\natexlab{a}})Wang, Zhao, Yatskar, Chang, and
  Ordonez]{balanced}
Tianlu Wang, Jieyu Zhao, Mark Yatskar, Kai-Wei Chang, and Vicente Ordonez.
\newblock Balanced datasets are not enough: Estimating and mitigating gender
  bias in deep image representations.
\newblock In \emph{The IEEE International Conference on Computer Vision
  (ICCV)}, October 2019{\natexlab{a}}.

\bibitem[Wang et~al.(2019{\natexlab{b}})Wang, Gonzalez-Garcia, van~de Weijer,
  and Herranz]{fair1}
Yaxing Wang, Abel Gonzalez-Garcia, Joost van~de Weijer, and Luis Herranz.
\newblock Controlling biases and diversity in diverse image-to-image
  translation.
\newblock \emph{arXiv preprint arXiv:1907.09754}, 2019{\natexlab{b}}.

\bibitem[Wilson et~al.(2019)Wilson, Hoffman, and Morgenstern]{unfair_od2}
Benjamin Wilson, Judy Hoffman, and Jamie Morgenstern.
\newblock Predictive inequity in object detection.
\newblock In \emph{Proceedings of the IEEE Conference on Computer Vision and
  Pattern Recognition Workshops}, 2019.

\bibitem[Xie et~al.(2017)Xie, Girshick, Doll{\'a}r, Tu, and He]{resnext}
Saining Xie, Ross Girshick, Piotr Doll{\'a}r, Zhuowen Tu, and Kaiming He.
\newblock Aggregated residual transformations for deep neural networks.
\newblock In \emph{Proceedings of the IEEE conference on computer vision and
  pattern recognition}, pages 1492--1500, 2017.

\bibitem[Zafar et~al.(2017)Zafar, Valera, Gomez~Rodriguez, and Gummadi]{odds}
Muhammad~Bilal Zafar, Isabel Valera, Manuel Gomez~Rodriguez, and Krishna~P
  Gummadi.
\newblock Fairness beyond disparate treatment \& disparate impact: Learning
  classification without disparate mistreatment.
\newblock In \emph{Proceedings of the 26th international conference on world
  wide web}, pages 1171--1180, 2017.

\bibitem[Zhang et~al.(2019)Zhang, Ren, Qiu, and Li]{inpainting1}
Ruonan Zhang, Yurui Ren, Jingfei Qiu, and Ge~Li.
\newblock Base-detail image inpainting.
\newblock In \emph{British Machine Vision Conference (BMVC)}, Cardiff, UK,
  2019.

\bibitem[Zhang et~al.(2017)Zhang, Song, and Qi]{utk}
Zhifei Zhang, Yang Song, and Hairong Qi.
\newblock Age progression/regression by conditional adversarial autoencoder.
\newblock In \emph{Proceedings of the IEEE conference on computer vision and
  pattern recognition}, pages 5810--5818, 2017.

\bibitem[Zhao et~al.(2017)Zhao, Wang, Yatskar, Ordonez, and Chang]{manalso}
Jieyu Zhao, Tianlu Wang, Mark Yatskar, Vicente Ordonez, and Kai-Wei Chang.
\newblock Men also like shopping: Reducing gender bias amplification using
  corpus-level constraints.
\newblock In \emph{Proceedings of the 2017 Conference on Empirical Methods in
  Natural Language Processing}, pages 2941--2951, 2017.
\newblock URL \url{https://www.aclweb.org/anthology/D17-1319}.

\bibitem[Zheng et~al.(2019)Zheng, Yu, Liu, and Zhang]{derain}
Yupei Zheng, Xin Yu, Miaomiao Liu, and Shunli Zhang.
\newblock Residual multiscale based single image deraining.
\newblock In \emph{British Machine Vision Conference (BMVC)}, Cardiff, UK,
  2019.

\bibitem[Zheng and Sun(2019)]{cvpr_vae}
Zhilin Zheng and Li~Sun.
\newblock Disentangling latent space for vae by label relevant/irrelevant
  dimensions.
\newblock In \emph{Proceedings of the IEEE Conference on Computer Vision and
  Pattern Recognition}, pages 12192--12201, 2019.

\bibitem[Zhu et~al.(2017)Zhu, Park, Isola, and Efros]{cycle}
Jun-Yan Zhu, Taesung Park, Phillip Isola, and Alexei~A Efros.
\newblock Unpaired image-to-image translation using cycle-consistent
  adversarial networks.
\newblock In \emph{Proceedings of the IEEE international conference on computer
  vision}, pages 2223--2232, 2017.

\end{thebibliography}
\end{document}